# EvaluationNet: Can Human Skill be Evaluated by Deep Networks?

Seong Tae Kim

Yong Man Ro

School of Electrical Engineering, KAIST, Republic of Korea
`{stkim4978, ymro}@kaist.ac.kr`

**Abstract**

*With the recent substantial growth of media such as YouTube, a considerable number of instructional videos covering a wide variety of tasks are available online. Therefore, online instructional videos have become a rich resource for humans to learn everyday skills. In order to improve the effectiveness of the learning with instructional video, observation and evaluation of the activity are required. However, it is difficult to observe and evaluate every activity steps by expert. In this study, a novel deep learning framework which targets human activity evaluation for learning from instructional video has been proposed. In order to deal with the inherent variability of activities, we propose to model activity as a structured process. First, action units are encoded from dense trajectories with LSTM network. The variable-length action unit features are then evaluated by a Siamese LSTM network. By the comparative experiments on public dataset, the effectiveness of the proposed method has been demonstrated.*

## 1. Introduction

As the recent substantial growth on media such as YouTube, a considerable number of instructional videos covering a wide variety of tasks are available online. Therefore, online instructional videos have become a rich resource for humans to learn everyday skills [1]. From everyday skills such as cooking or sports to specialized skills such as medical surgery or art, instructional video could be used to develop learner's competence. In order to improve the effectiveness of the learning with instructional video, observation and evaluation of the activity by experts are required. However, it is difficult for experts to observe and evaluate every activity steps. Recently, deep learning technology has been dramatically advanced and applied to various applications such as image classification [2-4], biometric authentication [5-8], medical image analysis [9-11], and action recognition [12-16]. At this point of view, it is natural to have a question that "Is it possible for deep networks to evaluate human skill?" In our scenario, users record their activity (i.e. human skill) and the deep network evaluates the user's activity compared with instructional video which is recorded from expert. For this purpose, evaluating two videos in the terms of semantic concept is important. In other words, calculating the semantic similarity of two videos is important. However, due to the reason that human activity's inherent variability, it is challenging task to evaluate the user's activity compared with the instructional video.

Most of the works on action recognition [12-22] have typically relied on unstructured models of video sequences. A holistic visual representation is usually computed over an entire video clip and then passed to a discriminative classifier to yield a single categorization label per video. These methods have been successful for the recognition of single action video clips. However, they do not appear to be well suited for the recognition of activities that require the modeling of complex behavior sequences.

Research in cognitive psychology has shown that human perceive as hierarchical structures rather than as flat [23]. In our scenario, it is possible to segment a continuous video into distinct meaningful events (i.e. action unit) by indicating the break point when user follows the instructional video. In order to deal with the inherent variability of activities, we propose to model activity as a structured process. Overall, the contributions of this study are mainly in followings:

1) We propose a deep network which encodes action units for the analysis of temporal structures. In this framework, we incorporate improved dense trajectory features and Fisher Vectors (FVs) into deep learning framework. Visual features of individual video frames are extracted and encoded by improved dense trajectory features and FVs. Action units are modeled by a long short-term memory (LSTM) network on the FV representation. The LSTM network could encode the motion of human and state change of the objects from the variable length of sequences. The action unit features are learned by the LSTM network to encode the underlying meaning expressed in a short sequence.

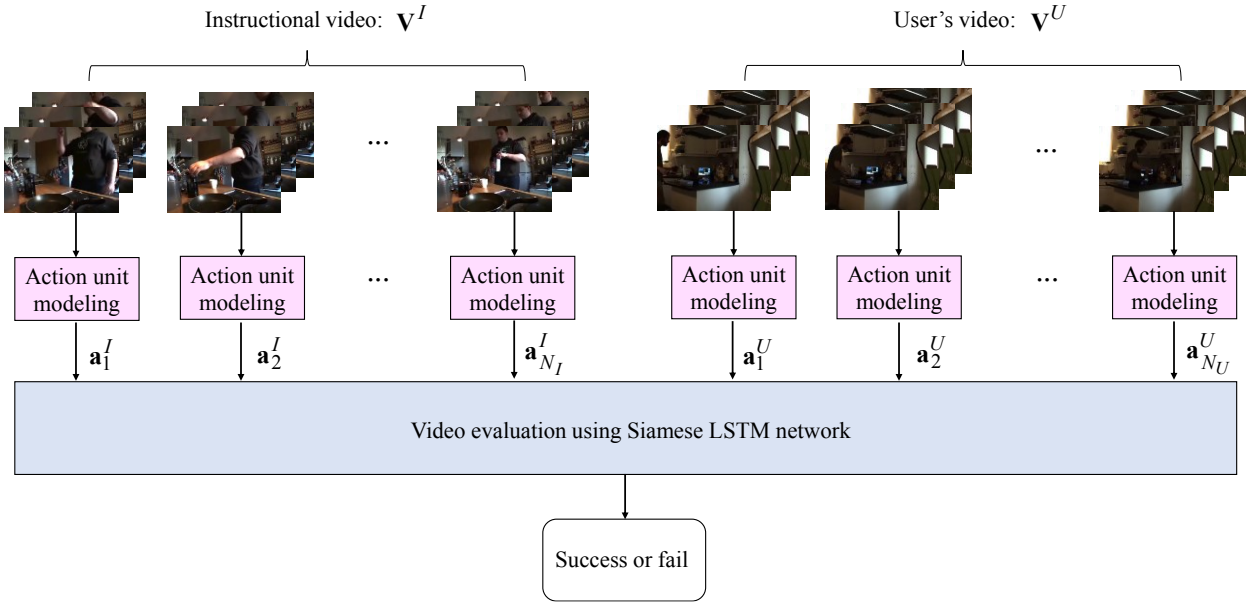

Figure 1: Overall procedure of the proposed method for video evaluation.

2) We propose a novel deep network for evaluating user's video compared with the instructional video in terms of semantic similarity. In order to encode relationships of the action units and assess semantic similarity between videos which comprised of variable-length action units, the LSTM network has been designed in Siamese structure. We compel the video representations learned by our model to form a highly structured space which reflects complex semantic relationships. By the comparative experiments on public dataset, the effectiveness of the proposed method has been verified for cooking activity. The proposed method could encode action units more accurately compared with the conventional methods. Moreover, by effectively evaluating semantic similarity of two videos consist of variable number of action units, the proposed method achieved more accurate evaluation results compared with other metric learning methods.

The rest of the paper is organized as follows. The proposed video evaluation method based on action unit modeling is explained in section 2. Comparative experiments and results are presented in section 3. The conclusions are drawn in section 4.

## 2. Proposed deep network for vided evaluation

Overall procedure of the proposed video evaluation method is shown in Figure 1. The aim of the proposed method is to evaluate the test video (i.e. user's video which records user's activity) compared with the instructional video. In our scenario, it is assumed that the users segment a continuous video into distinct meaningful events (i.e. action unit) by indicating the break point. In order to model complex behavior sequences, the proposed method has been devised as a structured process. In other words, the proposed method consists of action unit modeling and video evaluation using Siamese LSTM network. The details are described in the following subsections.

### 2.1. Action unit modeling

In this subsection, an action unit modeling using visual features and a LSTM network is described. The procedure of action unit modeling is shown in Figure 2. Action recognition researches in the last decade has explored on local features [17, 18]. Many action recognition literatures included sampling spatio-temporal descriptors [17, 18] and aggregating the descriptors by using VLAD [21, 22] or FVs [19] for video representation. In particular, dense trajectory features combined with FV aggregation showed outstanding results [20]. Improved version of dense trajectory features [3, 4] with FV aggregation was utilized for visual feature extraction by following the procedure described in [5]. In other words, the improved version of the dense trajectory features were extracted from each frames of video and the dimensionality of the feature was reduced by PCA and FV aggregation.

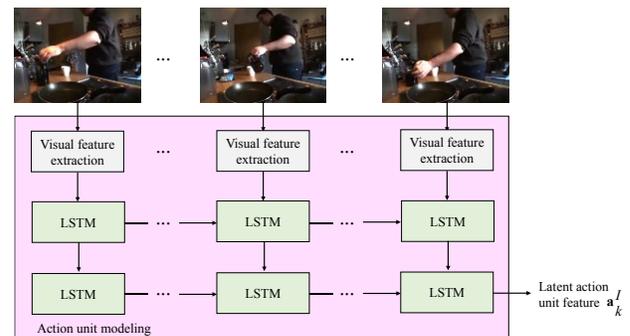

Figure 2: Action unit modeling using visual feature extraction and LSTM network.

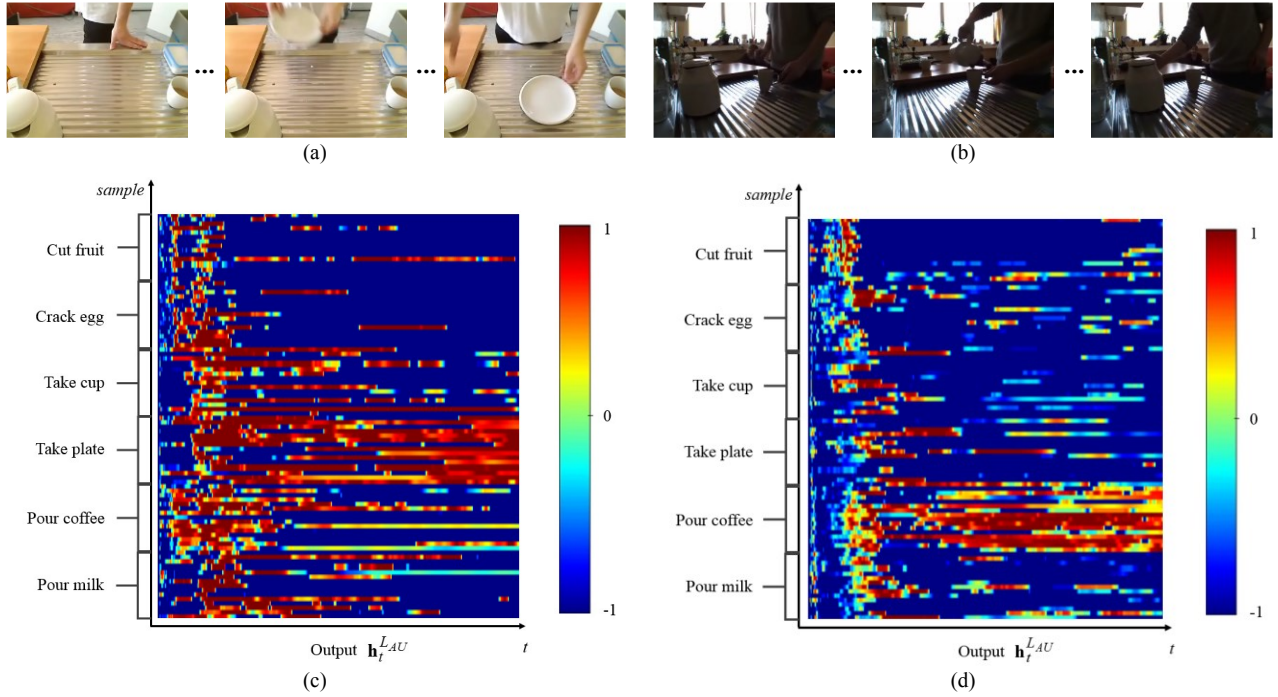

Figure 3: Visualization of an output for LSTM cell. For visualization purpose, the hidden state of action units of cut fruit, crack egg, take cup, take plate, pour coffee, and pour milk were selected. (a) Example for action unit of take plate. (b) Example for action unit of pour coffee. (c) Output of a LSTM cell which learns patterns of take plate. (d) Output of a LSTM cell which learns patterns of pour coffee. Best viewed in color.

The relationship of visual features extracted from each frame is learned by the LSTM network. The LSTM network incorporates memory cells with three control gates (i.e. input, forget, output) to learn long-term dependencies [24]. The memory cells can store, modify, and access internal state, which enable the network to discover latent action unit features. Moreover, the variable length sequence could be processed by the LSTM network.

In this paper, we make our LSTM network deep over temporal dimension, which has temporal recurrence of hidden variables. The deep LSTM network is constructed by stacking multiple LSTM layers on top of each other [8, 25]. Specifically, output from the lower LSTM layer ($(l-1)$-th layer), is used as an input of the upper LSTM layer ($l$-th later) as follows:

$$\mathbf{h}_t^l = \begin{cases} LSTM_{AU}(\mathbf{h}_{t-1}^l, \mathbf{f}_t), & if\ l = 1 \\ LSTM_{AU}(\mathbf{h}_{t-1}^l, \mathbf{h}_t^{l-1}), & otherwiese, \end{cases} \quad (1)$$

where $\mathbf{h}_t^l$ denotes hidden state at $t$-th time step of $l$-th stacked layer where $t = \{1,...,T\}$ and $l = \{1,...,L_{AU}\}$. $T$ denotes the number of sampled frames and $L_{AU}$ denotes the number of stacked LSTM layers in the LSTM network for action unit modeling. $\mathbf{f}_t$ denotes visual feature extracted from $t$-th sampled frame. $LSTM_{AU}(\cdot)$ denotes the function that performs the operations of the LSTM layer in the LSTM network for action unit modeling. This stacked LSTM network could combine the multiple representations with flexible use of action unit sequences [16]. In this paper, two stacked LSTM layers are used (i.e. $L_{AU} = 2$). The feature extracted from last time step $\mathbf{h}_T^{L_{AU}}$ is used as an action unit feature $\mathbf{a}_k^I$ where $k = 1,2,\cdots,N_I$. $N_I$ denotes the number of action unit of the instructional video. The dimension of the LSTM output vector $\mathbf{h}_T^{L_{AU}}$ is set to 128 (i.e. each LSTM layer has 128 memory cells).

The objective function ($E_I$) for learning the LSTM network for action unit modeling is devised as:

$$E_1 = -\sum_j \log \hat{p}_j, \quad (2)$$

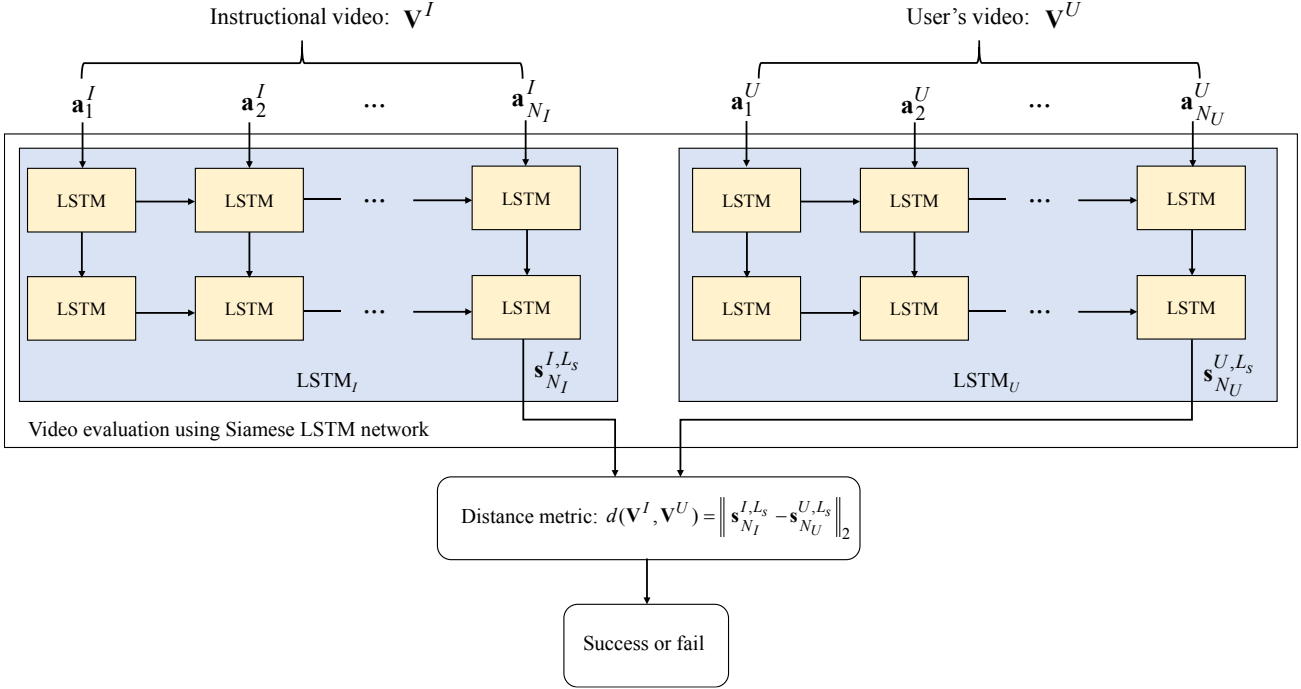

Figure 4: Structure of Siamese LSTM network for video evaluation.

where $\hat{p}_j$ denotes the probability which *j*-th training sample belongs to the true action unit class.

Figure 3 visualizes the evolution of an output for LSTM cell. The output $\mathbf{h}_t^{L_{AU}}$ denotes the latent action unit feature representation. As the network processes more frames, the memory cell states gradually extract the useful information related to corresponding action unit.

## 2.2 Video evaluation using Siamese LSTM network

In order to evaluate the user's video compared with the instructional video, the proposed Siamese LSTM model is designed as shown in Figure 4. There are two networks $LSTM_I$ and $LSTM_U$ which process a set of the action unit vectors in a given pair. By sharing weights of two networks $LSTM_I$ and $LSTM_U$ in the Siamese structure, we could effectively learn the deep network. The LSTM learns a mapping from the space of variable-length action unit feature vectors $\mathbf{A}^I = [\mathbf{a}_1^I, \mathbf{a}_2^I, \cdots, \mathbf{a}_{N_I}^I]$ into activity vector $\mathbf{s}_{N_I}^{I,L_s} \in \mathbf{R}^{d_{rep}}$ where $N_I$ and $L_s$ denote the number of action unit vectors in the instructional video and the depth of the stacked LSTM layers in the Siamese LSTM network, respectively. More concretely, each action unit vector is passed to the LSTM layer, which updates its hidden state at each step. Due to the reason that the number of action units could be changed in each test video, the Siamese LSTM network is designed to deal with the variable length of action unit features. The final representation of the activity of the video is encoded by $\mathbf{s}_{N_I}^{I,L_s} \in \mathbf{R}^{d_{rep}}$, which is the last hidden state of the model. We make our LSTM network deep over temporal dimension and two stacked LSTM layers are designed. The dimension of the LSTM output vector $d_{ref}$ is set to 128 (i.e. each LSTM layer has 128 memory cells in the Siamese LSTM network).

The objective function ($E_2$) for learning the Siamese LSTM network is devised based on contrastive loss [26]:

$$E_2 = \sum_i [y_i \cdot D_i + (1 - y_i) \cdot \{\max(0, m - D_i)\}^2], \quad (3)$$

where $y_i$ denotes the label of the user's video which is 1 for success video and 0 for fail video. $D_i = \left\| \mathbf{s}_{N_I}^{I,L_s} - \mathbf{s}_{N_U}^{U,L_s} \right\|_2$ denotes the distance of activity feature of instructional video $\mathbf{V}^I$ and activity feature of user's video $\mathbf{V}^U$.

By training the Siamese LSTM network using the objective function, the network could evaluate the test video whether it successfully conduct activities compared with the instructional video.

Table 1: Action units for individual activities.

| Activity | Action units |
|---|---|
| Coffee | take cup - pour coffee - pour milk - pour sugar - spoon sugar - stir coffee |
| Milk | take cup - spoon powder - pour milk - stir milk |
| Juice | take squeezer - take glass - take plate - take knife - cut orange - squeeze orange - pour juice |
| Tea | take cup - add teabag - pour water - spoon sugar - pour sugar - stir tea |
| Cereals | take bowl - pour cereals - pour milk - stir cereals |
| Fried Egg | pour oil - butter pan - take egg - crack egg - fry egg - take plate - add salt and pepper - put egg onto plate |
| Pancakes | take bowl - crack egg - spoon flour - pour flour - pour milk - stir dough - pour oil - butter pan - pour dough into pan - fry pancake - take plate - put pancake onto plate |
| Salad | take plate - take knife - peel fruit - cut fruit - take bowl - put fruit to bowl - stir fruit |
| Sandwich | take plate - take knife - cut bun - take butter - smear butter - take topping - add topping - put bun together |
| Scrambled Egg | pour oil - butter pan - take bowl - crack egg - stir egg - pour egg into pan - stir fry egg - add salt and pepper - take plate - put egg onto plate |

## 3. Experiments

### 3.1. Dataset

To evaluate the feasibility of the proposed approach, we have conducted comparative experiments on breakfast dataset which is a large dataset of daily cooking activities [23, 27, 28]. It had 10 activities of breakfast preparation such as making coffee, orange juice, chocolate milk, tea, cereals, fried eggs, pancakes, salad, sandwich, and scrambled eggs. Activities were performed by 52 different individuals in 18 different kitchen environments. The recording setup was "in the wild" in order to closely reflect real-world conditions as it pertains to the monitoring and analysis of daily activities. The actors were only handed a recipe and were instructed to prepare the corresponding food items. Due to the reason that the sequence were recorded in various kitchens, the subjects used the tools and packages that were locally available. Examples of the various settings and viewpoints are shown in Figure 5. A total of 1,712 video clips were annotated in action units. Some activities included shared action units (e.g. take cup or glass, pour milk or juice). This yield a low inter-class variance for activities combined with a high intra-class variance because of different recording locations, view-points and kitchens. Table 1 shows the action units corresponding to individual activities in the dataset. For evaluation purpose, the videos recorded from 52 subjects were divided to four groups as guided in the database [23, 27, 28]. Four fold cross-validation was conducted to report the performance.

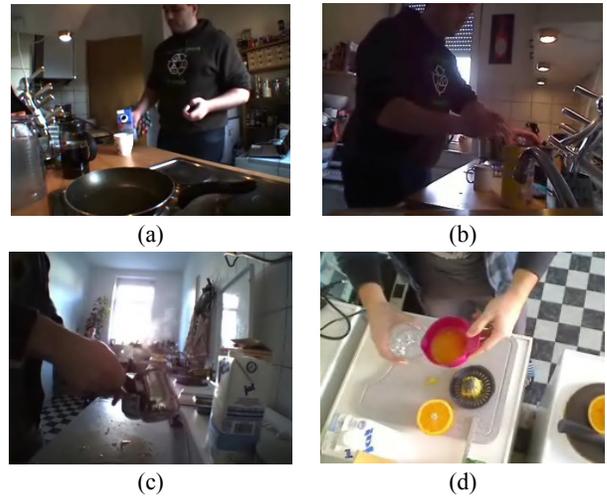

Figure 5: Example images from the breakfast dataset. (a) Image for action unit of pour milk from coffee activity of subject P03. (b) Image for action unit of spoon powder from milk activity of subject P03. (c) Image for action unit of pour coffee from coffee activity of subject P26. (d) Image for action unit of pour juice from juice activity of subject P26.

Table 2: Action unit classification on the Breakfast dataset based on coarse labels with 48 classes (1,712clips).

| Method | Accuracy |
|---|---|
| Proposed method | 37.6% |
| HMM [23] | 31.8% |
| SVM [28] | 21.8% |

Table 3: Activity evaluation results for the Breakfast dataset.

| Method | AUC |
|---|---|
| Proposed method | 0.886 |
| KISS metric learning [29] | 0.831 |
| SVM [30] | 0.798 |
| Fractional power cosine similarity [31,32] | 0.772 |

### 3.2. Action unit classification

We first evaluated the performance of the proposed action unit modeling method for the classification of individual action units. In other words, the performance of classification of pre-segmented videos into 48 action unit classes were measured. We compared the classification accuracy of the proposed method with HMM [23] and a linear SVM [28] using the same feature representation. For the case of the linear SVM, the FV representation was computed for the entire segment instead of each frame. The proposed method and the HMMs modeled the temporal relation of the observations. The result of action unit classification is summarized in Table 2. As shown in the table, SVM method achieved accuracy of 21.8% which was lower than other methods. Due to the reason that the SVM method did not consider the temporal relation of the observations, the accuracy of action unit classification was low. The HMM method achieved accuracy of 31.8%. The proposed method achieved accuracy of 37.6% which was the highest accuracy. It was mainly attributed to the fact that the proposed method effectively encoded the latent action unit features from the motion of human and state change of the objects by using the LSTM network.

### 3.3. Activity evaluation

To evaluate the performance of activity evaluation compared with different metric learning methods, all pairs of each fold were evaluated by setting one video as instructional video and the other video as test video. If the instructional video and the test video were same activity, the test video was considered as success video. If not, the test video was considered as fail video. In order to validate the effectiveness of the proposed method, KISS (keep it simple and straightforward) metric learning [29], support vector machine (SVM) [30], and fractional power cosine similarity [31, 32] were compared. For a fair comparison with other methods, we trained the other methods on the same action unit features. For the case of other feature similarity metrics, we performed average pooling on action unit features to get the fixed-length video descriptor. For calculating similarity, L2 normalization was conducted for each video-level feature vectors. As shown in Figure 6, the proposed method achieved higher true positive rates (TPR) over all false positive rates (FPR). The area under ROC curves (AUC) was also calculated and reported in Table 3. The AUC values of KISS metric learning, SVM, and Fractional power cosine similarity were 0.831, 0.798, and 0.772, respectively. The proposed method achieved AUC value of 0.886 and outperformed other evaluation methods.

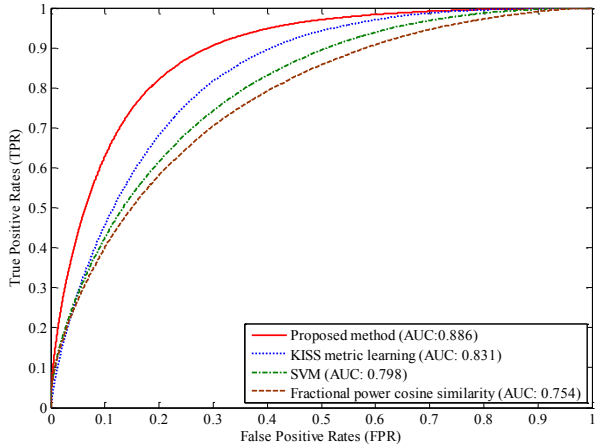

Figure 6: Video evaluation results for the Breakfast dataset. ROC curves obtained from different evaluation methods are plotted.

### 4. Conclusions

Online instructional videos have become a rich resource for humans to learn everyday skills. In order to improve the effectiveness of the learning with instructional video, evaluation of the activity are required. In this study, the novel deep learning framework which targeted human activity evaluation for learning from instructional video was proposed. In order to deal with the inherent variability of activities, we proposed to model activity as a structured process. First, action units were encoded from dense trajectories and FV aggregation with LSTM network. The variable-length action unit features were then evaluated by the Siamese LSTM network. By the comparative experiments on public dataset, the effectiveness of the proposed method was verified.

This study worth in the view point of raising new research problem. In other words, it is expected that further researches will be explored to solve the automatic evaluation problem for applications such as instructional video based education and personal robotic tasks. As a future work, we have a plan to extend our method with automatic temporal action detection framework for improving the flexibility and practicability of the proposed video evaluation framework.